\documentclass[runningheads]{llncs}
\usepackage{graphicx}

\usepackage{comment}
\usepackage{amsmath,amssymb} 
\usepackage{color}

\usepackage{multirow}
\usepackage{multicol}

\usepackage[accsupp]{axessibility}  

\usepackage[pagebackref,breaklinks,colorlinks]{hyperref}

\begin{document}
\pagestyle{headings}
\mainmatter
\def\ECCVSubNumber{1312}  

\title{Cross Attention Based Style Distribution for Controllable Person Image Synthesis} 

\titlerunning{CASD for Controllable Person Image Synthesis}
%
\author{$\text{Xinyue Zhou}$\inst{1} \and
$\text{Mingyu Yin}$\inst{1} \and
$\text{Xinyuan Chen}$\inst{3} \and
$\text{Li Sun}$\inst{1,2}\thanks{Corresponding author, email: sunli@ee.ecnu.edu.cn. } \and
$\text{Changxin Gao}$\inst{4} \and
$\text{Qingli Li}$\inst{1}}
\authorrunning{$\text{X. Zhou}$ et al.}
%
\institute{$^1$Shanghai Key Laboratory of Multidimensional Information Processing, \\
$^2$Key Laboratory of Advanced Theory and Application in Statistics and Data Science,\\
East China Normal University, Shanghai, China \\
$^3$Shanghai AI Laboratory, Shanghai, China \\
$^4$Huazhong University of Science and Technology, Wuhan, China}


\maketitle

\begin{figure}
\centering
\includegraphics[width=1\textwidth]{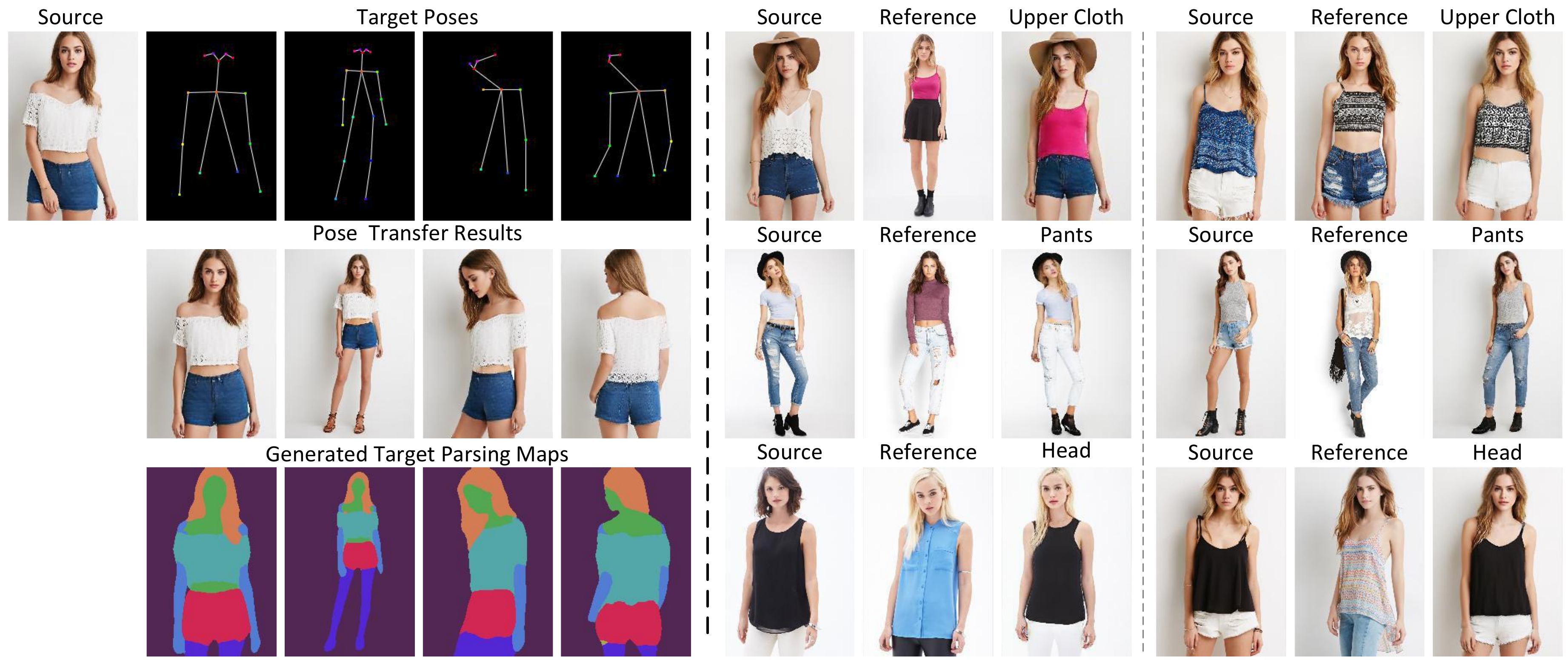}
\caption{Left: given the source image and target pose, our model is able to transfer the pose and generate the target parsing map as required. Note that we have only a single training stage without independent generation for the target parsing map. However, our model still synthesizes it precisely by cross attention based style distribution module.
Right: Our model also enables virtual try-on and head(identity) swapping by explicitly controlling the poses and per-body-part appearance of source and reference images.}
\label{fig:head_img}
\end{figure}

\begin{abstract}
Controllable person image synthesis task enables a wide range of applications through explicit control over body pose and appearance. 
In this paper, we propose a cross attention based style distribution module that computes between the source 
semantic styles and target pose for pose transfer.
The module intentionally selects the style represented by each semantic 
and distributes them according to the 
target pose. 
The attention matrix in cross attention expresses the dynamic similarities between the target pose and the source styles for all semantics. Therefore, it can be utilized to route the color and texture from the source image, and is further constrained by the target parsing map to achieve a clearer objective. 
At the same time, to encode the source
appearance accurately, the self attention among different semantic styles is also added.
The effectiveness of our model is validated quantitatively and qualitatively on pose transfer and virtual try-on tasks. Codes are available at \href{https://github.com/xyzhouo/CASD}{https://github.com/xyzhouo/CASD}.

\end{abstract}

\section{Introduction}\label{sec:intro}
Synthesizing realistic person images under explicit control of the body pose and appearance 
has many potential applications, such as person reID \cite{zheng2017unlabeled,wei2018person,ge2018fd}, video generation \cite{yang2018pose,liu2019liquid} and virtual clothes try-on \cite{han2018viton,wang2018toward,dong2019fw,ge2021parser}, \emph{etc.} 
Recently, 
the conditional GAN is employed to 
transfer the source style into the specified target pose. The generator 
connects the intended style with the required pose in its different layers. 
\emph{E.g.}, PATN \cite{zhu2019progressive}, HPT \cite{yang2021towards}, ADGAN \cite{men2020controllable} insert several repeated modules with the same structure to combine style and pose features. However, these modules are usually composed of common operations, such as Squeeze-and-Excitation (SE) \cite{hu2018squeeze} or Adaptive Instance Normalization (AdaIN) \cite{huang2017arbitrary}, which lacks the ability to align source style with target pose.

In contrast, the 2D or 3D deformation is applied in the task with a clearer motivation. 
DefGAN \cite{siarohin2018deformable}, GFLA \cite{ren2020deep} and Intr-Flow \cite{li2019dense} estimate the correspondence between the source and target pose to guide the spread of appearance features. 
Although these methods generate realistic texture, they may produce noticeable artifacts when faced with large deformations.
Besides, more than one training stages are often needed, and the unreliable flow from the first stage limits the quality of results. 

This paper aims for the better fusion on features of both source image and target pose in a single training stage. Instead of directly estimating the geometry deformation and warping source features to fulfill target pose, we propose a simple cross attention based style distribution (CASD) module to calculate between  the target pose and the source style represented  by each semantic, 
and distribute the source semantic styles to the target pose. 
The basic idea is to employ the 
coarse fusion features under target pose 
as queries, requiring the source styles from different semantic components as keys and values to update and refine them. 
Following ADGAN \cite{men2020controllable}, appearance within each semantic region is described by a style encoder, which extracts the color and texture 
within the corresponding region 
(such as head, arms, or legs, \emph{etc.}). The 
style features are dynamically 
distributed by the CASD block for each query position under the target pose. Particularly, values from each semantic are softly weighted and summed together according to the attention matrix, so that they 
are matched with target pose. The aligned feature can be further utilized to affect the 
input to the decoder.

To further improve the synthesis quality, we have some special designs within CASD block. First, to tightly link styles from different semantics, 
the self attention is performed among them, 
making each style no longer independent with others. 
Second, another routing scheme, in the same size with attention matrix, is also employed for style routing. It is directly predicted from the target pose without exhaustive comparisons with keys of styles. Third, extra constraint from target parsing map 
is incorporated on the attention matrix, so that the attention head has a clearer motivation. In this way, our attention matrix represents the predicted target parsing map. Additionally, our model can also achieve  virtual try-on and head(identity) swapping based on reference images by exchanging the specific semantic region in style features. Fig \ref{fig:head_img} shows some applications of our model.
The contributions of the paper can be summarized into following aspects.
 \begin{itemize}
 \item We propose cross attention based style distribution (CASD) module for controllable person image synthesis, which softly selects the source style represented by each semantic and distributes them to the target pose. 
 \item We intentionally add self attention to connect styles from different semantic components, and let the model predict the attention matrix based on the target pose. Moreover, the target parsing maps are used as the ground truths for the attention matrix, giving the model an evident object during training.
 \item We can achieve applications in image manipulation by explicit controlling over body pose and appearance, \emph{e.g.}, pose transfer, parsing  map generation, virtual try-on and head(identity) swapping. 
 \item Extensive experiments on 
 DeepFashion dataset validates the effectiveness of our proposed model. Particularly, the synthesis quality has been greatly improved, indicated by both quantitative metrics and user study.
 \end{itemize}
\section{Related Work}
\textbf{Human pose transfer} is first proposed in \cite{ma2017pose}, and becomes well developed in recent years due to the advancement in image synthesis. Most of the existing works need paired training data, which employ the ground truth under target pose during training. Though a few of them are fully unsupervised \cite{ma2018disentangled,esser2018variational,pumarola2018unsupervised,song2019unsupervised,zhang2020cross,zhang2020cross,zhou2021cocosnet,sanyal2021learning}, they are not of our major concern. Previous research can be characterized into either two- (or multi-) stage 
or one-stage methods. 
The former first generates coarse images or foreground masks, and then gives them to the second stage generator as input for refinement. 
In \cite{balakrishnan2018synthesizing}, 
the model first segments the foreground from image into different body components, and then applies learnable spatial deformations on them to generate the foreground image. 
GFLA \cite{ren2020deep} pretrains a network to estimate the 2D flow and occlusion mask based on source image, source and target poses. Afterwards, it uses them to warp local patches of the source 
to match the required pose. Li \emph{et.al.} \cite{li2019dense} fit a 3D mesh human model onto the 2D image, and train the first stage model to predict the 3D flow, 
which is employed to warp the source appearance in the second stage. LiquidGAN \cite{liu2019liquid} also adopts the 3D model to guide the geometry deformation 
within the foreground region. Although  geometry-based methods generate realistic texture, they may fail to extract accurate motions, resulting in noticeable artifacts.
On the other hand, without any deformation operation, PISE \cite{zhang2021pise} and SPGnet \cite{lv2021learning} synthesize the target parsing maps, given the source masks, source poses and target poses as input in the first stage. Then it generates the image with the help of them in the second stage. These work show that the parsing maps under target pose have potential to be exploited 
for pose transfer. 

Compared to two-stage methods, the single stage model has light training burden. Different from \cite{balakrishnan2018synthesizing}, DefGAN \cite{siarohin2018deformable} 
explicitly 
computes the local 2D affine transformation between source and target patches, and applies the deformation 
to align source features to the target pose. PATN \cite{zhu2019progressive} proposes a repeated pose attention module, consisting of the computation like SE-Net, to combine features from the appearance and pose. ADGAN \cite{men2020controllable} uses a texture encoder to extract style vectors within each semantic, and gives them to several AdaIN residual blocks to synthesize the target image. XingGAN \cite{tang2020xinggan} proposes two types of cross attention blocks to fuse features from the target pose and source appearance in two directions, repeatedly. Although these models design the fusion block of pose and appearance style, they lack the operation to align source appearance with the target pose. CoCosNet \cite{zhang2020cross,zhou2021cocosnet} computes the dense correspondences between cross-domain images
by attention-based operation. 
However, each target position is only related to a local patch of the source image, which implies that the correlation matrix should be a sparse matrix, and the dense correlation matrix leads to quadratic memory consumption. Our model deals with this problem by an efficient CASD block.

\textbf{Attention and transformer modules} first appear in NLP \cite{vaswani2017attention}, which enlarge the receptive field in a dynamic way. Non-local network \cite{wang2018non} is its first attempt in image domain. The scheme becomes increasingly popular in various tasks including image classification \cite{dosovitskiy2020image,touvron2021training,liu2021swin,wang2021pyramid}, object detection \cite{carion2020end,zhu2020deformable} and semantic segmentation \cite{huang2019ccnet,zheng2021rethinking} due to its effectiveness. There are basically two different ways for it which are self and cross attention. Self attention projects the queries, keys and values from the same token set, while cross attention usually obtains keys and values from one set, and queries from another one. The computation process then becomes the same, measuring the similarity between queries and keys to form an attention matrix, which is used to weight values to update query tokens. Based on the repeated attention module, multi-stage transformer can be built. Note that adding the MLP (FFN) and residual connection between stages are crucial and become a designing routine, for which we also follows.


\section{Method}\label{sec:sec3}
\subsection{Overview Framework}
\begin{figure*}[t]
\centering
\includegraphics[width=1\textwidth]{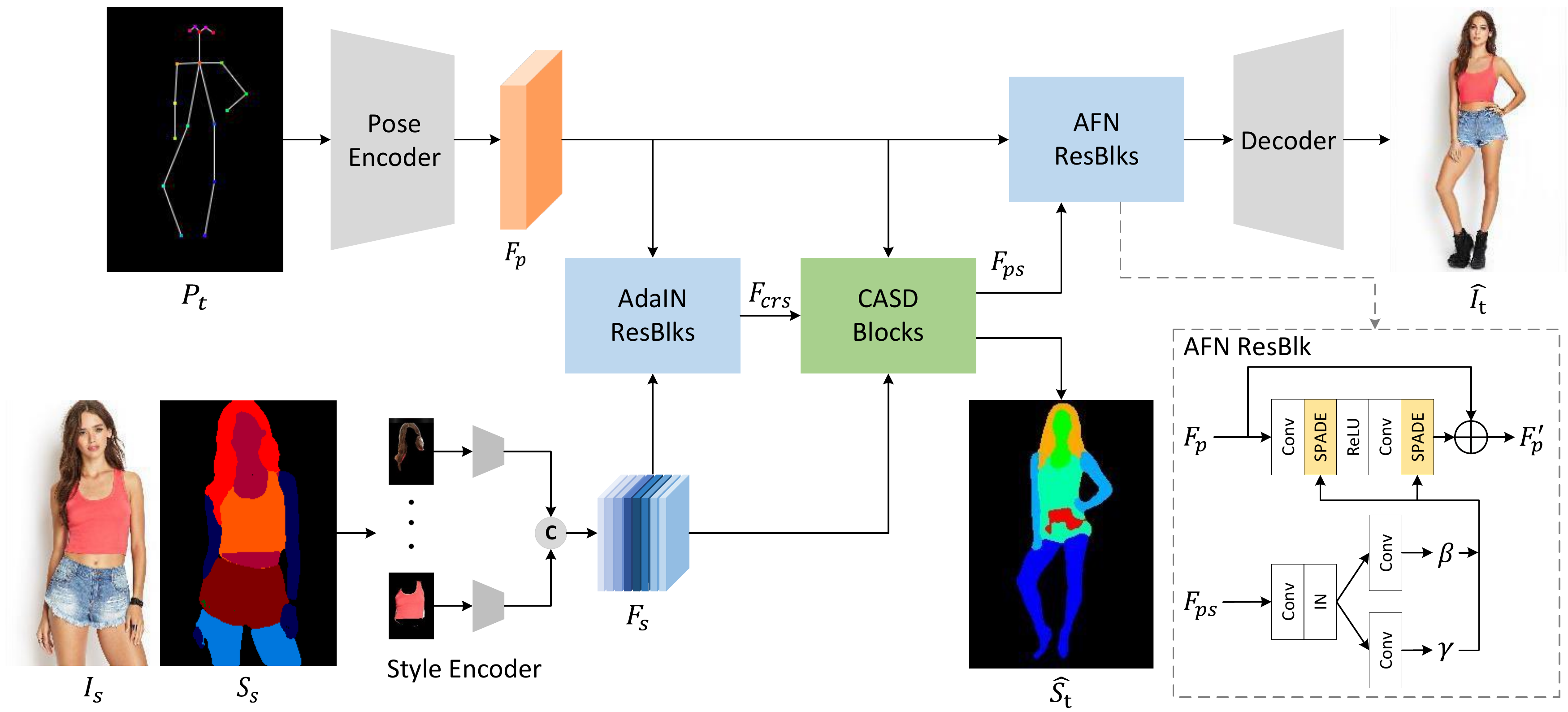}
\caption{Overview architecture of our proposed generator. There are separate pose and style encoders, with their outputs $F_p$ and $F_s$ being fused by AdaIN ResBlks and Cross Attention based Style Distribution (CASD) blocks, and giving the pose-aligned feature $F_{ps}$ as the output. Then, the same $F_{ps}$ is adapted to 
decoder through AFN ResBlks. The key component, CASD Block, consists of both self and cross attention, and can also output the predicted target parsing maps $\hat{S}_t$.}

\label{fig:fig1}
\end{figure*}

Given a source image $I_s$ under the pose $P_s$, our goal is to synthesize a high fidelity image $\hat{I}_t$ under a different target pose $P_t$. 
$\hat{I}_t$ should not only fulfil the pose requirement,  
but also have the same appearance with $I_s$. Fig \ref{fig:fig1} shows the overview of the proposed generation model. It consists of a pose encoder $E_p$, a semantic region style 
encoder $E_s$ and a decoder $Dec$. Besides, there are several cross attention based style distribution (CASD) blocks which are the key components in our generator. Before and after the attention, there are several AdaIN residual blocks and Aligned Feature Normalization (AFN) residual blocks with the similar design as \cite{park2019semantic,yin2020novel}. The former coarsely adapts the source style to the target pose, while the latter incorporates the pose-aligned feature from the CASD blocks into the decoder. Both of them 
are learnable, 
which change the feature statistics in their affecting layers. 

As is shown in Fig \ref{fig:fig1}, the desired $P_t$ is directly used as the input by encoder $E_p$, which describes the key point positions of human body. For each point, we make a single channel heatmap with a predefined standard deviation to describe its location. Except the individual point, we additionally adopt straight lines between selected points 
to better model the pose structure.
There are totally $18$ points and $12$ lines, so $P_t \in \mathbb{R}^{H\times W\times 30}$. To facilitate accurate style 
extraction from $I_s$, we follow the strategy in \cite{men2020controllable}, which employs the source parsing map $S_s\in\mathbb{R}^{H\times W\times N_s}$ to separate the full image into regions, so that $E_s$ independently encodes the styles in different semantics. $N_s$ is the total number of semantics in the parsing map. During training, the ground truth image $I_t$ and its corresponding parsing map $S_t$ are exploited. 
Note that $F_p\in \mathbb{R}^{H\times W\times C}$ and $F_s\in\mathbb{R}^{N_s\times1\times1\times C}$ represent the pose and style features from $E_p$ and $E_s$, respectively. They are utilized by the CASD blocks, whose details are introduced in the following section. Moreover, the CASD block is repeated by multiple times, \emph{e.g.} twice, gradually completing the fusion between $F_p$ and $F_s$, and forming a better aligned feature $F_{ps}$. Finally, $F_{ps}$ is given to AFN 
from the side branch to take its effect. Besides, our CASD block can also output the predicted target parsing map $\hat{S}_t$ by constraining the cross attention matrix.

\subsection{Pre- and Post-Attention Style injection}
Since the source style $F_s^i\in\mathbb{R}^{1\times 1\times C}$ is independently encoded by a shared-weight encoder $E_s$, where $i=1,2,\cdots,N_s$ is the semantic index, 
they may not appropriate for 
style injection together. 
So we first 
combine 
$F_s^i$ from different semantic regions through an MLP, 
and give results to AdaIN ResBlks to roughly combine $F_s$ with $F_p$, specifying the coarse fusion $F_{crs}$ reflecting the target pose $P_t$, which then 
participates cross attention as queries in the CASD blocks.

After the CASD blocks, we have $F_{ps}$ which is obviously superior to $F_{crs}$, 
and is suitable for the utilization by $Dec$. Instead of giving $F_{ps}$ directly to $Dec$, we design an AFN ResBlks to employ it as a conditional feature. Within the block, an offset $\beta$ and a scaling factor $\gamma$ are first predicted. Then, they take effect through AFN, which performs the conditional normalization according to $\beta$ and $\gamma$. Note that $F_{ps}$, $\beta$ and $\gamma$ are in the same size.

\begin{figure}[t]
\centering
\includegraphics[width=0.5\columnwidth]{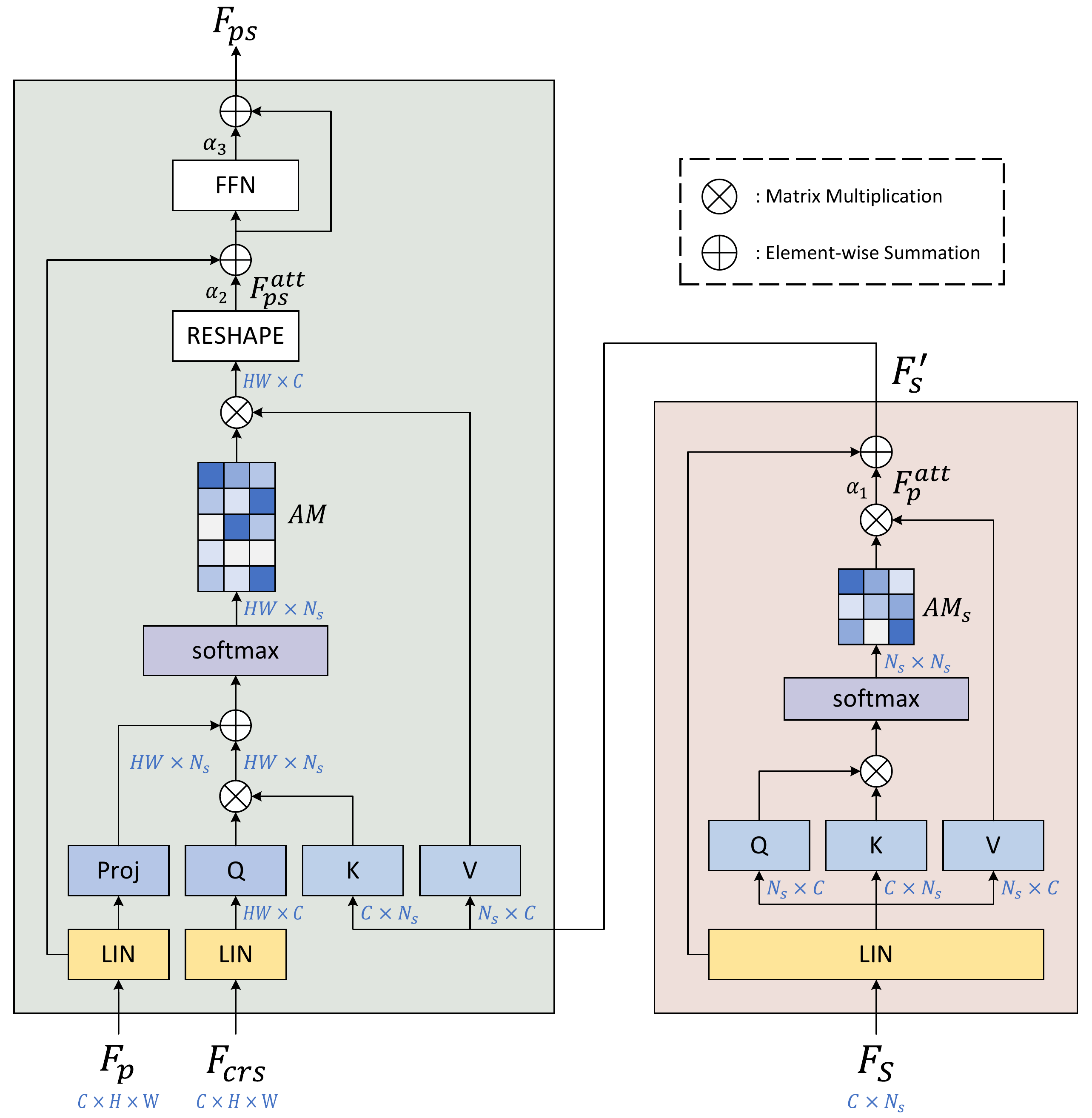}
\caption{Illustrations of the cross attention based style distribution (CASD) block. On the right, self attention is performed on the source style features $F_s$, giving the updated $F_s'$ as the output. On the left, cross attention is computed between the coarsely aligned feature $F_{crs}$ and the updated style feature $F_s'$. The pose $F_p$ also joins the cross attention. And by constraining the cross attention matrix $AM$, predicted target parsing maps $\hat{S}_t$ is generated. Note that $\alpha_{1-3}$ indicates the learnable scaling factors. 
}
\label{fig:query}
\end{figure}

\subsection{Cross Attention based Style Distribution Block}
The computation in CASD block includes two stages, which are self attention 
and cross attention stages, as is depicted in Fig \ref{fig:query}. The two types of attention are carried out in sequence, and they together align the source style $F_s$ according to the target pose feature $F_p$. We describe them in the following two sections. 

\subsubsection{Self attention for Style Features}
The self attention is performed among $F_s^i$, so that each 
$F_s^i$ of a particular semantic is connected with others $F_s^j$ where $j\ne i$. We simplify the classic design on self attention module in transformer \cite{vaswani2017attention}. Traditionally, there are three learnable projection heads $W_Q$, $W_K$ and $W_V$, and they are in the same shape $W_Q,W_K,W_V\in\mathbb{R}^{C\times C}$. These heads are responsible for mapping the input tokens $F_s$ into the query $Q$, key $K$ and value $V$. In our application, to reduce learnable parameters, we omit $W_Q$ and $W_K$, and directly use $F_s$ as both query and key. However, we keep $W_V$ and it yields $V \in \mathbb{R}^{N_s\times C}$ in the same dimension of $F_s$. The self attention, computing the update style feature $F_s^{att}$, can be summarized into Eq (\ref{eq:eq1}).
\begin{equation}\label{eq:eq1}
\begin{aligned}
F_s^{att} & =\text{Attention}(Q,K,V)=\text{Softmax}(QK^{\text{T}}/\sqrt{C})V \\
 Q & =F_s,\quad K=F_s, \quad V=F_s W_V
\end{aligned}
\end{equation}

Here the attention module essentially compares the similarities of different semantic styles, so that each $F_s^i$ in $F_s$ absorbs information from 
other style tokens $F_s^j$. Note that we also follow the common designs in transformer. Particularly, there is a residual connection between $F_s$ and $F_s^{att}$, so they are added, and given to later layers. 
The final style 
is denoted by $F_s'$, as is shown on the right of Fig \ref{fig:query}. Moreover, different from traditional transformer, we employ Layer Instance Normalization (SW-LIN) proposed in \cite{xu2021drb} to replace LN for better synthesis. 

\subsubsection{Cross Attention}
The cross attention module further adapts the source style 
$F_s'$ into the required pose, shown on the left of Fig \ref{fig:query}. 
Such attention is carried out across different domains, between the 
coarse fusion $F_{crs}$ output from AdaIN resblks and the style feature $F_s'$. Therefore, different from the previous self attention, its result $F_{ps}$ has spatial dimensions, and actually reflects how to distribute the source style under an intended pose. In this module, $F_{crs}$ from AdaIN ResBlks is treated as queries $Q$. Specifically, there are totally $H\times W$ unique queries. Each of them is a $C$-dim vector. $F_s'$ provides the keys $K$ for comparison with $Q$ and values $V$ for soft selection. 

Here we aggregate the common attention computation as is shown in Eq (\ref{eq:eq2}). $F_{ps}^{att}\in\mathbb{R}^{H\times W\times C}$ is the updated amount on query after the attention, in the same size with $F_{crs}$. 
\begin{equation}\label{eq:eq2}
\begin{aligned}
F_{ps}^{att}& =\text{Attention}(Q,F_p,K,V)=AM \cdot V \\ 
& =\left(\text{Softmax}\left(\frac{QK^{\text{T}}}{\sqrt{C}} + \text{Proj}(F_p)\right)\right)V 
\end{aligned}
\end{equation}
Note that we set $Q=F_{crs}W_Q$, $K=F_s'W_K$ and $V=F_s'W_V$, so the attention actually combines the 
features 
$F_{crs}$ with source styles $F_s'$. In Eq (\ref{eq:eq2}), the first term in the bracket indicates the regular attention matrix which exhaustively computes the similarity between every $Q$-$K$ pair. It is of the shape $H\times W\times N_s$ and determines the possible belonging semantic for each 
position. Since there are 
projection heads $W_Q$ and $W_K$ to adjust query and key, the attention matrix is fully dynamic, which is harmful to model convergence. Our solution is to add the second term with the same shape as the first one, forming the augmented attention matrix $AM$, and let it 
participate value routing. $\text{Proj}(\cdot)$ is a linear projection head which directly outputs a routing scheme based only on $F_p$. Therefore, it implies that the model is able to predict the target parsing map for each position, given the encoded pose feature $F_p$. Some recent works like SPGNet \cite{lv2021learning} or PISE \cite{zhang2021pise} has a separate training stage to generate the target parsing map according to the required pose. Our model has a similar intention, but it is more convenient with only a single training stage. In the following section, we add a constraint on the attention matrix to generate the predicted target parsing map.

After the cross attention in Eq (\ref{eq:eq2}), we follow the routine in transformer, which first makes the element-wise summation between $F_{ps}^{att}$ and $F_p$, then gives the result to an $\text{FFN}$, leading to a better pose feature $F_{ps}$ which combines the source style for the next stage.

\subsection{Learning Objectives}
Similar to the previous method \cite{zhu2019progressive,men2020controllable}, we employ the adversarial loss $L_{adv}$, reconstruction loss $L_{rec}$, perceptual loss $L_{perc}$ and contextual loss $L_{CX}$ as our learning objectives. Additionally, we also adopt an attention matrix cross-entropy loss $L_{AMCE}$ and an LPIPS loss $L_{LPIPS}$ to train our model. The full learning objectives are formulated in Eq (\ref{eq:eq9}), 
\begin{equation}\label{eq:eq9}
\begin{aligned}
L_{full} = \lambda_{adv} L_{adv} + \lambda_{rec} L_{rec} + \lambda_{perc} L_{perc}+  \lambda_{CX} L_{CX} \\+\lambda_{AMCE} L_{AMCE} + \lambda_{LPIPS} L_{LPIPS}
\end{aligned}
\end{equation}
where $\lambda_{adv}$, $\lambda_{rec}$, $\lambda_{perc}$, $\lambda_{CX}$,  $\lambda_{AMCE}$ and $\lambda_{LPIPS}$ are hyper-parameters controlling the relative importance of these objectives. They are detailed as follows.

\noindent\textbf{Attention Matrix Cross-entropy Loss. } To train our model with an evident object, we adopt cross-entropy loss to constrain the attention matrix $AM$ close to target parsing map $S_t$, which is defined as:
\begin{equation}\label{eq:eq3}
\begin{aligned}
L_{AMCE} = -\sum_{i=1}^H\sum_{j=1}^W\sum_{c=1}^{N_s}S_t\left(i, j, c\right)\log\left(AM\left(i, j, c\right)\right). 
\end{aligned}
\end{equation}
where $i$, $j$ denote the position of spatial dimension in the attention matrix $AM$, and $c$ denotes the position of semantic dimension in attention matrix $AM$. By employing this loss in the training process, our model can generate the predicted target parsing map in a single stage. 

\noindent\textbf{Adversarial loss. } We adopt a 
pose discriminator $D_p$ and a style discriminator $D_s$ to help 
$G$ generate more realistic result in adversarial training. Specifically, real pose pairs $({P_t, I_t})$ and fake pose pairs $({P_t, \hat{I}_t})$ are feed into $D_p$ for pose consistency. Meanwhile, real image pairs $({I_s, I_t})$ and fake image pairs $({I_s, \hat{I}_t})$ are feed into $D_s$ for style consistency. Note that both discriminators are trained with $G$ in an end-to-end way to promote each other.
\begin{equation}\label{eq:eq4}
\begin{aligned}
L_{adv} = \mathbb{E}_{I_s, I_t, P_t} \left[ \log \left(D_s \left(I_s, I_t\right) \cdot D_p \left(P_t, I_t\right) \right)\right] \\ + 
\mathbb{E}_{I_s, P_t} [ \log \left(1 - D_s \left(I_s, G \left(I_s, P_t \right)\right)\right) \\
\cdot  \left(1 - D_p \left(P_t, G \left(I_s, P_t\right) \right)\right)]
\end{aligned}
\end{equation}
\noindent\textbf{Reconstruction and perceptual loss.} The reconstruction loss $L_{rec}$ is used to encourage the generated image $\hat{I}_t$ to be similar with ground-truth $I_t$ at the pixel level, 
which is computed as
$L_{rec} = \parallel \hat I_t - I_t \parallel_1$.
The perceptual loss 
calculates the $L_1$ distance between the features extracted from the pre-trained VGG-19 network \cite{simonyan2014very}. It can be written as
$L_{perc} = \sum_{i} \parallel \phi_i \left(\hat{I}_t\right) -  \phi_i \left({I}_t\right) \parallel_1$,
where $\phi_i$ is the feature map of the i-th layer of the pre-trained VGG-19 network.

\noindent\textbf{Contextual Loss.} 
We also adopt contextual loss, which is first proposed in \cite{mechrez2018contextual}, aiming to measure the similarity between two non-aligned images for image transformation. It is computed in Eq (\ref{eq:eq7}).
\begin{equation}\label{eq:eq7}
\begin{aligned}
L_{CX} = - \log \left(CX\left( F^l\left(\hat I_t\right), F^l \left(I_t\right)\right)\right)
\end{aligned}
\end{equation}
Here $F^l(\hat I_t)$ and $F^l(I_t)$ denotes the feature extracted from layer $l = relu\{3\_2, 4\_2\}$ of the pre-trained VGG-19 
for images $\hat I_t$ and $I_t$, respectively, and CX denotes the cosine similarity metric between 
features.

\noindent\textbf{LPIPS Loss.} 
In order to reduce distortions and 
learn perceptual similarities, we integrate the LPIPS loss \cite{zhang2018unreasonable}, which has been shown to better preserve image quality compared to the more standard perceptual loss:
\begin{equation}\label{eq:eq8}
\begin{aligned}
L_{LPIPS} = \parallel F \left( \hat I_t \right)  - F \left( I_t \right)  \parallel_2
\end{aligned}
\end{equation}
where $F(\cdot)$ denotes the perceptual feature extracted from pre-trained VGG-16 network. 

\section{Experiments}


\subsection{Experimental Setup}

\noindent\textbf{Dataset.}
We carry out experiments on DeepFashion (In-shop Clothes Retrieval Benchmark) \cite{liuLQWTcvpr16DeepFashion},
which contains $52,712$ high-quality person images with the resolution of $256\times 256$. 
Following the same data configuration in \cite{zhu2019progressive}, we split this dataset into training and testing subsets with 101,966 and 8,570 pairs, respectively. Additionally, we use the segmentation masks 
obtained from the human parser \cite{gong2017look}. 
Note that the person 
ID of the training and testing sets do not overlap.

\noindent\textbf{Evaluation Metrics.} 
We employ four metrics SSIM \cite{wang2004image}, FID \cite{heusel2017gans}, LPIPS \cite{zhang2018unreasonable} and PSNR  
for evaluation. Peak Signal to Noise Ratio (PSNR) and Structural Similarity Index Measure (SSIM) is the most commonly used 
in 
image generation task with known ground truths. The former utilizes the mean square error to give an overall evaluation, while the latter 
calculates the global variance and mean 
to assess the structural similarity. 
Meanwhile, Learned Perceptual Image Patch Similarity (LPIPS) 
is another metric to compute the distance between the generations and ground truths 
in the perceptual domain. 
Besides, Fréchet Inception Distance (FID) is employed to measure the realism of the generated images. It calculates the Wasserstein-2 distance between the distributions of the generated and real data. 

\noindent\textbf{Implementation Details. } Our method is implemented in PyTorch and trained 2 NVIDIA Tesla-A100 GPUs with the batch size 
being equal to 16. We adopt Adam optimizer \cite{kingma2014adam} with $\beta_1 = 0.5$, $\beta_2 = 0.999$ to train our model for around 330k iterations, using the same epochs with other works \cite{men2020controllable}. 
The weights of the learning objectives are set as: $\lambda_{AMCE} = 0.1$, $\lambda_{LPIPS} = 1$, $\lambda_{rec} = 1$, $\lambda_{perc} = 1$, $\lambda_{adv} = 5$ and $\lambda_{CX} = 0.1$, without tuning. The number of semantic part is $N_s=8$, which includes the ordinary semantics as background, pants, hair, glove, face, dress, arms and legs. Furthermore, the learning rate is initially set to 0.001, and linearly decayed to 0 after 115k iterations.
Following above configuration, we alternatively optimize the generator and two discriminators.
We train our model for pose transfer task as described in Sec \ref{sec:sec3}, after the convergence of training, we use the same trained model for all tasks, \emph{e.g.}, pose transfer, virtual try-on, head(identity) swapping and parsing map generation.


\begin{figure*}[t]
\centering
\includegraphics[width=0.6\textwidth]{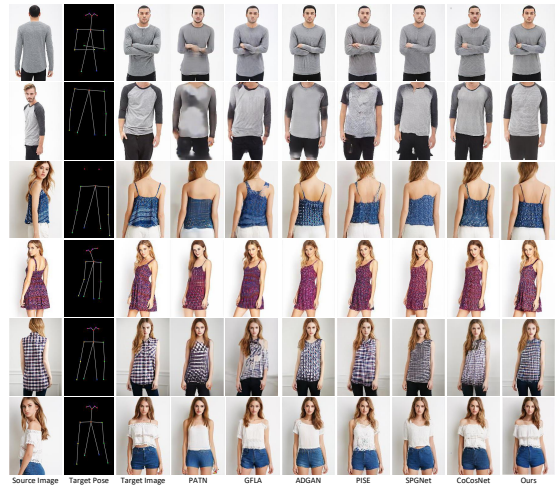}
\caption{Qualitative comparison between our method and other state-of-the-arts. The target ground truths and the synthesized results from each models are listed in rows. }
\label{fig:fig4}
\end{figure*}

\begin{table*}[ht]
\centering
\caption{Comparisons on metrics for image quality and user study. SSIM, FID, LPIPS and PSNR are the quantitative metrics for synthesized images. R2G, G2R and Jab are metrics computed from users' feedback.}\label{tab:tab1}
\setlength{\tabcolsep}{1.7mm}
\renewcommand\arraystretch{1.1}
\begin{tabular}{cccccccc}
\hline

\textbf{Models} & \textbf{SSIM$\uparrow$}  & \textbf{FID$\downarrow$}    & \textbf{LPIPS$\downarrow$} &
\textbf{PSNR$\uparrow$} &
\textbf{R2G$\uparrow$}   & \textbf{G2R$\uparrow$}   & \textbf{Jab$\uparrow$}  \\ \hline
PATN \cite{zhu2019progressive}            & 0.6709          & 20.7509          & 0.2562          & 31.14          & 19.14          & 31.78          & 0.26\%        \\
GFLA \cite{ren2020deep}           & 0.7074          & \textbf{10.5730}          & 0.2341          & 31.42          & 19.53          & 35.07          & 13.72\%       \\
ADGAN \cite{men2020controllable}          & 0.6721          & 14.4580          & 0.2283          & 31.28          & 23.49 & 38.67          & 11.17\%       \\
PISE \cite{zhang2021pise}           & 0.6629          & 13.6100          & 0.2059 & 31.33          & -              & -              & 14.89\%       \\
SPGNet \cite{lv2021learning}         & 0.6770          & 12.2430 & 0.2105         & 31.22           & 19.47          & 36.80           & 17.26\%       \\
CoCosNet \cite{zhang2020cross}         & 0.6746          & 14.6742 & 0.2437         & 31.07           & -          & -           & 13.73\%       \\
Ours            & \textbf{0.7248} & 11.3732          & \textbf{0.1936} & \textbf{31.67}          & \textbf{24.67}         & \textbf{40.52} & \textbf{28.96\%} \\ \hline
\end{tabular}
\setlength{\abovecaptionskip}{0pt}
\setlength{\belowcaptionskip}{10pt}
\end{table*}

\subsection{Pose Transfer}
In this section, 
we compare our method with several state-of-the-art methods, including PATN \cite{zhu2019progressive} , GFLA \cite{ren2020deep}, ADGAN \cite{men2020controllable}, PISE \cite{zhang2021pise} , SPGNet \cite{lv2021learning} and CoCosNet \cite{zhang2020cross}. Quantitative and qualitative results as well as user study are conducted to verify the effectiveness of our method. 
All the results are obtained by directly using the source code and well-trained models published by their authors. Since CoCosNet uses a different train/test split, we directly uses its well-trained model on our test set.

\noindent\textbf{Quantitative comparison. }
The quantitative results 
are listed in Table \ref{tab:tab1}. Notably, our method achieves the best performance on most metrics compared with the other methods, which 
can be attributed to the proposed CASD block.

\noindent\textbf{Qualitative comparison. }
In Fig \ref{fig:fig4}, we compare the generated results from different methods. 
It can be observed that our method produces more realistic and reasonable results (\emph{e.g.}, the second, third and penultimate rows).
More importantly, our model can well retain the details 
from the source image (\emph{e.g.}, the fourth and last rows).
Moreover, even if target pose is complex (\emph{e.g.}, the first row), our method can still generate it precisely.

\noindent\textbf{User study. }  
While both quantitative and qualitative comparisons can evaluate the performance of the generated results in different aspects, human pose transfer tasks tend to be user-oriented.
Therefore, we conduct a user study with 30 volunteers to evaluate the performance in terms of human perception.
The user study consists of two parts.
$\left(i\right)$ Comparison with ground-truths. Following \cite{lv2021learning}, we randomly select 30 real images and 30 generated images from test set and shuffle them. Volunteers are required to determine whether a given image is real or fake within a second.  
$\left(ii\right)$ Comparison with the other methods, we present volunteers 30 random selected image pairs that include source image, target pose, ground-truth and images generated by our method and baselines. 
Volunteers are asked to select the most realistic and reasonable image with respect to the source image and ground truth.
Note that we shuffle all the generated images for fairness. The results are shown in the right part of Table \ref{tab:tab1}.  Here we adopt three metrics, namely \textbf{R2G}: the percentage of the real images treated as the generated images; \textbf{G2R}: the percentage of the generated images treated as real images;  \textbf{Jab}: the percentage of images judged to be the best among all models. Higher values of these three metrics mean better performance. 
We can observe that our model achieves the best results, especially about $11\%$ higher than the 2-nd best one on Jab. 

\begin{figure}[t]
\centering
\includegraphics[width=0.6\textwidth]{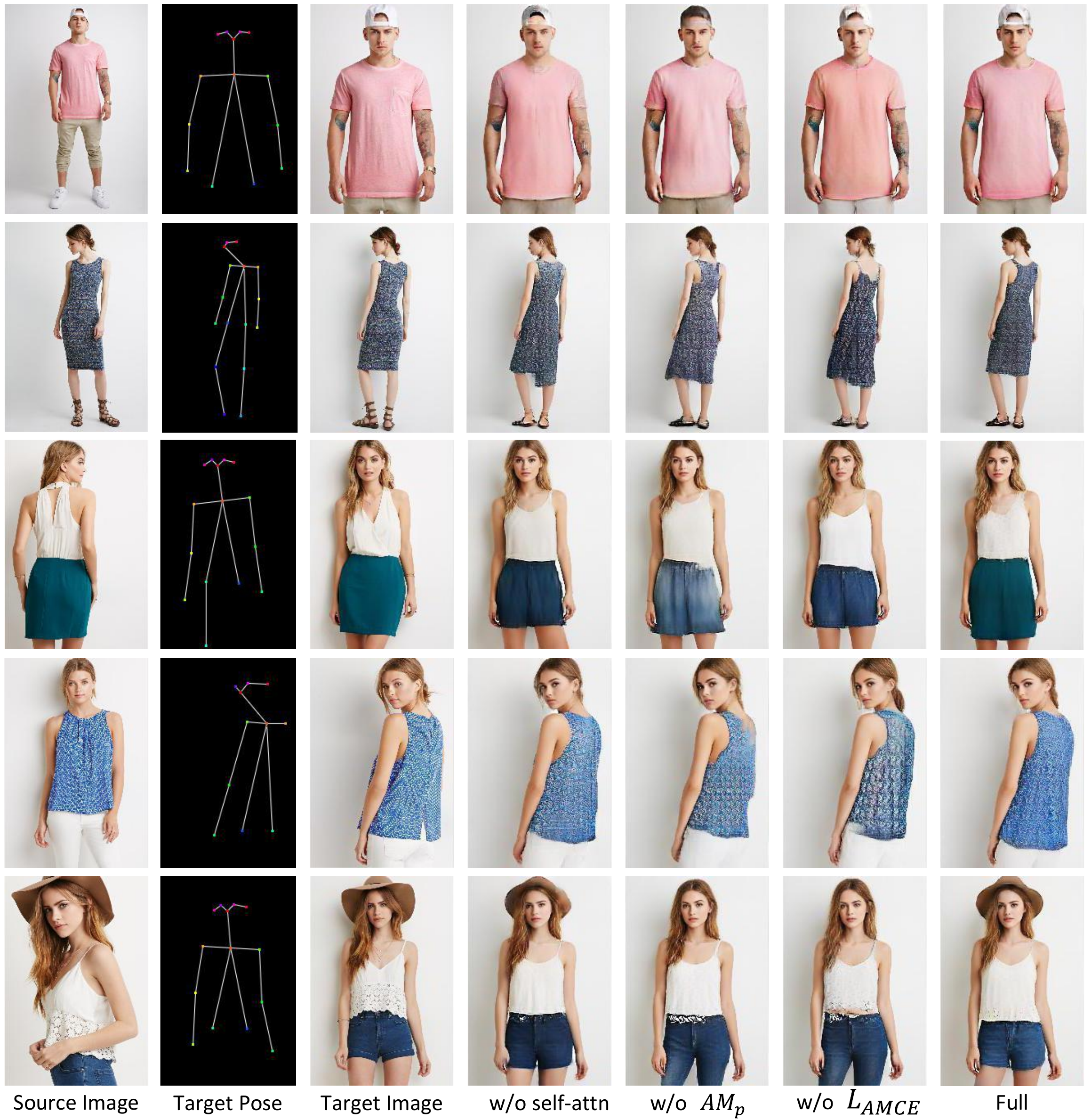}
\caption{The qualitative results of ablation study. The ground truths and the synthesize images from each ablation model are listed in columns.}
\label{fig:fig5}
\end{figure}

\begin{table}[h]
\centering
\caption{Quantitative ablations on each proposed component in the full model. The performances of the final model are given the last row. In the above three rows, we intentionally exclude one component from the full model. Details are given in Sec \ref{sec:sec43}.}
\setlength{\tabcolsep}{2.5mm}
\renewcommand\arraystretch{1.1}
\begin{tabular}{ccccc}
\hline
\textbf{Model} & \textbf{SSIM$\uparrow$}  & \textbf{FID$\downarrow$}    & \textbf{LPIPS$\downarrow$}  &
\textbf{PSNR$\uparrow$} \\ \hline
w/o self-attn  & 0.7201          & 13.2462          & 0.2017          & 31.52          \\
w/o $AM_p$        & 0.7213          & 12.4265          & 0.1985          & 31.48          \\
w/o $L_{AMCE}$       & 0.7156          & 14.7513          & 0.2126          & 31.41          \\
\hline
Ours-Full      & \textbf{0.7248} & \textbf{11.3732} & \textbf{0.1936} & \textbf{31.67} \\ \hline
\end{tabular}
\setlength{\abovecaptionskip}{0pt}
\setlength{\belowcaptionskip}{10pt}
\label{tab:tab2}
\end{table}

\begin{figure}[t]
\centering
\includegraphics[width=0.9\textwidth]{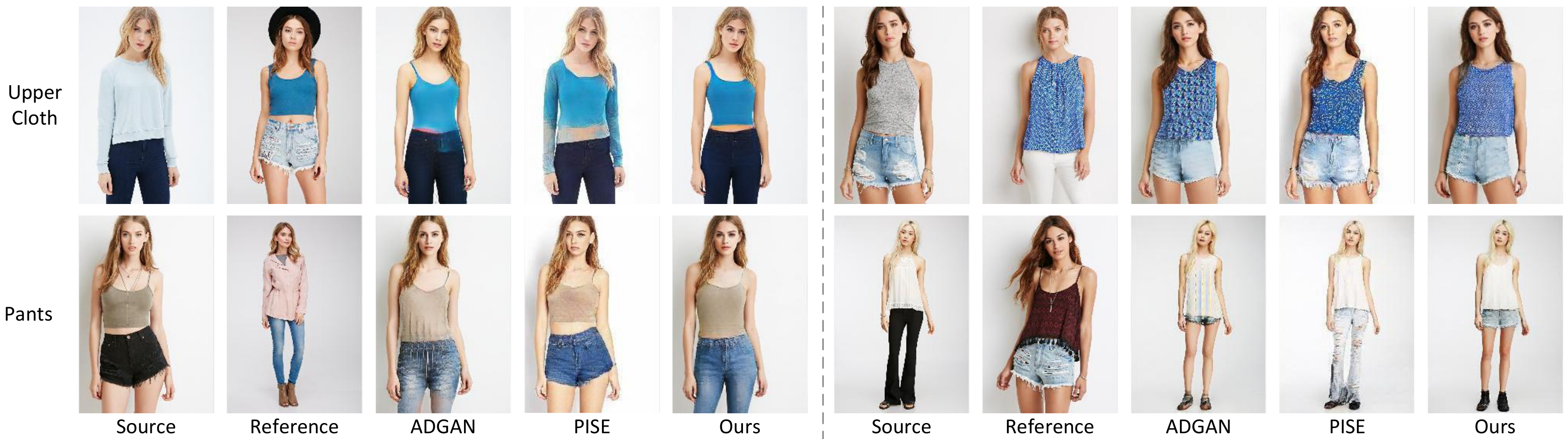}
\caption{The visual comparisons with other state-of-the-art methods on virtual try-on. }
\label{fig:fig6}
\end{figure}

\begin{table}[h]
\centering
\caption{Comparisons of the FID score and user study with other state-of-the-art methods on virtual try-on and head(identity) swapping tasks. }
\setlength{\tabcolsep}{2.5mm}
\renewcommand\arraystretch{1.1}
\begin{tabular}{ccccccc}
\hline
\multirow{2}{*}{\textbf{Method}} & \multicolumn{2}{c}{\textbf{Upper Cloth}} & \multicolumn{2}{c}{\textbf{Pants}} & \multicolumn{2}{c}{\textbf{Head}} \\ \cline{2-7} 
                        & \textbf{FID}$\downarrow$           & \textbf{Jab}$\uparrow$      & \textbf{FID}$\downarrow$              & \textbf{Jab}$\uparrow$          & \textbf{FID}$\downarrow$           & \textbf{Jab}$\uparrow$       \\ \hline
ADGAN                   & 14.3720       & 24.67\%        & 14.4446          & 29.67\%             & 14.4596       & 23.58\%          \\
PISE                    & 14.0537       & 22.92\%         & 14.2874          & 22.25\%             & 14.3647       & 30.75\%          \\ \hline
Ours                    & \textbf{12.5376}       & \textbf{52.41}\%          & \textbf{12.5456}          & \textbf{48.08}\%             & \textbf{12.6578}       & \textbf{45.67}\%          \\ \hline
\end{tabular}
\setlength{\abovecaptionskip}{0pt}
\setlength{\belowcaptionskip}{10pt}
\label{tab:tab3}
\end{table}

\subsection{Ablation Study}\label{sec:sec43}
In this section, we perform ablation study to further verify our assumptions and evaluate the contribution of each component in our model.
We implement 3 variants by alternatively removing a specific component from the full model (w/o self-attn, w/o$AM_p$, w/o $L_{AMCE}$). 


\noindent\textbf{W/o self-attn. } This model removes self attention in CASD blocks, only uses cross attention, which directly feeds $F_s$ into cross attention as Key and Value. 

\noindent\textbf{W/o $AM_p$. } The model removes $AM_p=\text{Proj}(Q)$ in CASD blocks in Eq (\ref{eq:eq2}), which will not let the model predict the attention matrix based on the target pose.

\noindent\textbf{W/o $L_{AMCE}$. } The model does not adopt $L_{AMCE}$ loss defined in Eq (\ref{eq:eq3}) for training, 
so it can not be explicitly guided by the target parsing map during cross attention.

\noindent\textbf{Full model. } It includes all components and achieves the best performance on all quantitative metrics, as is shown Table \ref{tab:tab2}. Meanwhile, it also gives the best visual results as is shown in Fig \ref{fig:fig5}. It’s shown that by removing any parts of our proposed model would lead to a performance drop.


\subsection{Virtual Try-on and Head Swapping}

Benefiting from the semantic region style encoder, 
our model can also achieve controllable person image synthesis based on reference images by exchanging the channel feature of specific semantic region in the style features (\emph{e.g.}, upper-body transfer, lower-body transfer and head swapping) without further training. We compare our method with ADGAN \cite{men2020controllable} and PISE \cite{zhang2021pise}. The visual comparisons are shown in Fig \ref{fig:fig6}. We observe that our model can reconstruct target part and retain other remaining parts more faithfully. In addition, when transferring the lower-body, PISE cannot transfer the target pants to the source person, it will retain the shape of the source person's pants and only transfer the texture. 

For more comprehensive comparisons, quantitative comparison and user study are also conducted. The results are shown in Table \ref{tab:tab3}. In the user study, we randomly select 40 results generated by our method and the other compared methods for each task, and then we invite 30 volunteers to select the most realistic results. \textbf{Jab} is the percentage of images judged to be the best among all methods.

\begin{table*}[]
\centering
\caption{Comparison of per-class IoU with SPGNet on the predicted target parsing maps.}\label{tab:tab4}
\setlength{\tabcolsep}{1mm}
\renewcommand\arraystretch{1.1}
\begin{tabular}{cccccccccc}
\hline
\textbf{Model} & \textbf{pants} & \textbf{hair}  & \textbf{gloves} & \textbf{face}  & \textbf{u-clothes} & \textbf{arms}  & \textbf{legs}  & \textbf{Bkg}   & \textbf{Avg}   \\ \hline
SPGNet \cite{lv2021learning}        & 42.18          & \textbf{66.51} & \textbf{9.36}   & 62.46          & 67.87              & 58.46          & 44.13          & 84.65          & 61.89          \\
Ours           & \textbf{49.02} & 65.87          & 5.50            & \textbf{67.24} & \textbf{76.72}     & \textbf{59.90} & \textbf{50.81} & \textbf{90.28} & \textbf{66.34} \\ \hline
\end{tabular}
\setlength{\abovecaptionskip}{0pt}
\setlength{\belowcaptionskip}{10pt}
\end{table*}

\subsection{Target parsing map synthesis}
Moreover, to intuitively understand of our CASD blocks, we further 
show the predicted target parsing maps in Fig \ref{fig:head_img}. It shows when given the source image and various target poses, our model can not only transfer the poses, but also synthesize the target parsing maps, though we do not separately build a model to do this. We list the Intersection over Union (IoU) metric between predictions from our method and \cite{gong2017look} for all semantics in Table \ref{tab:tab4}. For major semantics, we achieve higher IoU than SPGNet \cite{lv2021learning}. Note that our model gives the final synthesis images and target parsing map in one stage. The synthesized paired data can be used as training data for segmentation.

\begin{figure*}[t]
\centering
\includegraphics[width=0.8\textwidth]{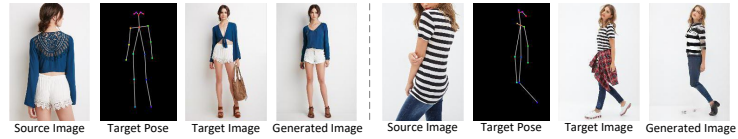}
\caption{Failure cases caused by incomprehensible garment (left) or pose (right). }
\setlength{\abovecaptionskip}{10pt}
\label{fig:fig7}
\end{figure*}

\section{Limitations}

Although our method produces impressive results in most cases, it still fails to generate incomprehensible garments and poses. As shown in Fig \ref{fig:fig7}, a specific knot on a blouse fails to generate and a person in a rare pose can not be synthesized seamlessly. We believe that training the model with more various images will alleviate this problem.

\section{Conclusion}
This paper presents a cross attention based style distribution block for a single-stage controllable person image synthesis task, 
which has strong ability to align the source semantic styles with the
target poses.
The cross attention based style distribution block mainly consists of self and cross attention, which not only captures the source semantic styles accurately, but also aligns them to the target pose precisely.
To achieve a clearer objective, the AMCE loss is proposed to constrain the attention matrix in cross attention by target parsing map. Extensive experiments and ablation studies show the satisfactory performance of our model, and the effectiveness of its components. Finally, we show that our model can be easily applied to virtual try-on and head(identity) swapping tasks.

\section*{Acknowledgements} 
This work is supported by the Science and Technology Commission of Shanghai Municipality No.19511120800, Natural Science Foundation of China No.61302125 and No.62102150, and ECNU Multifunctional Platform for Innovation(001).

\clearpage
%
%

\begin{thebibliography}{10}
\providecommand{\url}[1]{\texttt{#1}}
\providecommand{\urlprefix}{URL }
\providecommand{\doi}[1]{https://doi.org/#1}


\bibitem{balakrishnan2018synthesizing}
Balakrishnan, G., Zhao, A., Dalca, A.V., Durand, F., Guttag, J.: Synthesizing
  images of humans in unseen poses. In: Proceedings of the IEEE Conference on
  Computer Vision and Pattern Recognition. pp. 8340--8348 (2018)

\bibitem{carion2020end}
Carion, N., Massa, F., Synnaeve, G., Usunier, N., Kirillov, A., Zagoruyko, S.:
  End-to-end object detection with transformers. In: European Conference on
  Computer Vision. pp. 213--229. Springer (2020)

\bibitem{dong2019fw}
Dong, H., Liang, X., Shen, X., Wu, B., Chen, B.C., Yin, J.: Fw-gan:
  Flow-navigated warping gan for video virtual try-on. In: Proceedings of the
  IEEE/CVF International Conference on Computer Vision. pp. 1161--1170 (2019)

\bibitem{dosovitskiy2020image}
Dosovitskiy, A., Beyer, L., Kolesnikov, A., Weissenborn, D., Zhai, X.,
  Unterthiner, T., Dehghani, M., Minderer, M., Heigold, G., Gelly, S., et~al.:
  An image is worth 16x16 words: Transformers for image recognition at scale.
  arXiv preprint arXiv:2010.11929  (2020)

\bibitem{esser2018variational}
Esser, P., Sutter, E., Ommer, B.: A variational u-net for conditional
  appearance and shape generation. In: Proceedings of the IEEE Conference on
  Computer Vision and Pattern Recognition. pp. 8857--8866 (2018)

\bibitem{ge2018fd}
Ge, Y., Li, Z., Zhao, H., Yin, G., Yi, S., Wang, X., Li, H.: Fd-gan:
  Pose-guided feature distilling gan for robust person re-identification. arXiv
  preprint arXiv:1810.02936  (2018)

\bibitem{ge2021parser}
Ge, Y., Song, Y., Zhang, R., Ge, C., Liu, W., Luo, P.: Parser-free virtual
  try-on via distilling appearance flows. In: Proceedings of the IEEE/CVF
  Conference on Computer Vision and Pattern Recognition. pp. 8485--8493 (2021)

\bibitem{gong2017look}
Gong, K., Liang, X., Zhang, D., Shen, X., Lin, L.: Look into person:
  Self-supervised structure-sensitive learning and a new benchmark for human
  parsing. In: Proceedings of the IEEE Conference on Computer Vision and
  Pattern Recognition. pp. 932--940 (2017)

\bibitem{han2018viton}
Han, X., Wu, Z., Wu, Z., Yu, R., Davis, L.S.: Viton: An image-based virtual
  try-on network. In: Proceedings of the IEEE conference on computer vision and
  pattern recognition. pp. 7543--7552 (2018)

\bibitem{heusel2017gans}
Heusel, M., Ramsauer, H., Unterthiner, T., Nessler, B., Hochreiter, S.: Gans
  trained by a two time-scale update rule converge to a local nash equilibrium.
  Advances in neural information processing systems  \textbf{30} (2017)

\bibitem{hu2018squeeze}
Hu, J., Shen, L., Sun, G.: Squeeze-and-excitation networks. In: Proceedings of
  the IEEE conference on computer vision and pattern recognition. pp.
  7132--7141 (2018)

\bibitem{huang2017arbitrary}
Huang, X., Belongie, S.: Arbitrary style transfer in real-time with adaptive
  instance normalization. In: Proceedings of the IEEE International Conference
  on Computer Vision. pp. 1501--1510 (2017)

\bibitem{huang2019ccnet}
Huang, Z., Wang, X., Huang, L., Huang, C., Wei, Y., Liu, W.: Ccnet: Criss-cross
  attention for semantic segmentation. In: Proceedings of the IEEE/CVF
  International Conference on Computer Vision. pp. 603--612 (2019)

\bibitem{kingma2014adam}
Kingma, D.P., Ba, J.: Adam: A method for stochastic optimization. arXiv
  preprint arXiv:1412.6980  (2014)


\bibitem{li2019dense}
Li, Y., Huang, C., Loy, C.C.: Dense intrinsic appearance flow for human pose
  transfer. In: Proceedings of the IEEE/CVF Conference on Computer Vision and
  Pattern Recognition. pp. 3693--3702 (2019)

\bibitem{liu2019liquid}
Liu, W., Piao, Z., Min, J., Luo, W., Ma, L., Gao, S.: Liquid warping gan: A
  unified framework for human motion imitation, appearance transfer and novel
  view synthesis. In: Proceedings of the IEEE/CVF International Conference on
  Computer Vision. pp. 5904--5913 (2019)

\bibitem{liu2021swin}
Liu, Z., Lin, Y., Cao, Y., Hu, H., Wei, Y., Zhang, Z., Lin, S., Guo, B.: Swin
  transformer: Hierarchical vision transformer using shifted windows. arXiv
  preprint arXiv:2103.14030  (2021)

\bibitem{liuLQWTcvpr16DeepFashion}
Liu, Z., Luo, P., Qiu, S., Wang, X., Tang, X.: Deepfashion: Powering robust
  clothes recognition and retrieval with rich annotations. In: Proceedings of
  IEEE Conference on Computer Vision and Pattern Recognition (CVPR) (June 2016)

\bibitem{lv2021learning}
Lv, Z., Li, X., Li, X., Li, F., Lin, T., He, D., Zuo, W.: Learning semantic
  person image generation by region-adaptive normalization. In: Proceedings of
  the IEEE/CVF Conference on Computer Vision and Pattern Recognition. pp.
  10806--10815 (2021)

\bibitem{ma2017pose}
Ma, L., Jia, X., Sun, Q., Schiele, B., Tuytelaars, T., Van~Gool, L.: Pose
  guided person image generation. arXiv preprint arXiv:1705.09368  (2017)

\bibitem{ma2018disentangled}
Ma, L., Sun, Q., Georgoulis, S., Van~Gool, L., Schiele, B., Fritz, M.:
  Disentangled person image generation. In: Proceedings of the IEEE Conference
  on Computer Vision and Pattern Recognition. pp. 99--108 (2018)

\bibitem{mechrez2018contextual}
Mechrez, R., Talmi, I., Zelnik-Manor, L.: The contextual loss for image
  transformation with non-aligned data. In: Proceedings of the European
  Conference on Computer Vision (ECCV). pp. 768--783 (2018)

\bibitem{men2020controllable}
Men, Y., Mao, Y., Jiang, Y., Ma, W.Y., Lian, Z.: Controllable person image
  synthesis with attribute-decomposed gan. In: Proceedings of the IEEE/CVF
  Conference on Computer Vision and Pattern Recognition. pp. 5084--5093 (2020)

\bibitem{park2019semantic}
Park, T., Liu, M.Y., Wang, T.C., Zhu, J.Y.: Semantic image synthesis with
  spatially-adaptive normalization. In: Proceedings of the IEEE/CVF Conference
  on Computer Vision and Pattern Recognition. pp. 2337--2346 (2019)

\bibitem{pumarola2018unsupervised}
Pumarola, A., Agudo, A., Sanfeliu, A., Moreno-Noguer, F.: Unsupervised person
  image synthesis in arbitrary poses. In: Proceedings of the IEEE Conference on
  Computer Vision and Pattern Recognition. pp. 8620--8628 (2018)

\bibitem{ren2020deep}
Ren, Y., Yu, X., Chen, J., Li, T.H., Li, G.: Deep image spatial transformation
  for person image generation. In: Proceedings of the IEEE/CVF Conference on
  Computer Vision and Pattern Recognition. pp. 7690--7699 (2020)

\bibitem{sanyal2021learning}
Sanyal, S., Vorobiov, A., Bolkart, T., Loper, M., Mohler, B., Davis, L.S.,
  Romero, J., Black, M.J.: Learning realistic human reposing using cyclic
  self-supervision with 3d shape, pose, and appearance consistency. In:
  Proceedings of the IEEE/CVF International Conference on Computer Vision. pp.
  11138--11147 (2021)

\bibitem{siarohin2018deformable}
Siarohin, A., Sangineto, E., Lathuiliere, S., Sebe, N.: Deformable gans for
  pose-based human image generation. In: Proceedings of the IEEE Conference on
  Computer Vision and Pattern Recognition. pp. 3408--3416 (2018)

\bibitem{simonyan2014very}
Simonyan, K., Zisserman, A.: Very deep convolutional networks for large-scale
  image recognition. arXiv preprint arXiv:1409.1556  (2014)

\bibitem{song2019unsupervised}
Song, S., Zhang, W., Liu, J., Mei, T.: Unsupervised person image generation
  with semantic parsing transformation. In: Proceedings of the IEEE/CVF
  Conference on Computer Vision and Pattern Recognition. pp. 2357--2366 (2019)

\bibitem{tang2020xinggan}
Tang, H., Bai, S., Zhang, L., Torr, P.H., Sebe, N.: Xinggan for person image
  generation. In: European Conference on Computer Vision. pp. 717--734.
  Springer (2020)

\bibitem{touvron2021training}
Touvron, H., Cord, M., Douze, M., Massa, F., Sablayrolles, A., J{\'e}gou, H.:
  Training data-efficient image transformers \& distillation through attention.
  In: International Conference on Machine Learning. pp. 10347--10357. PMLR
  (2021)

\bibitem{vaswani2017attention}
Vaswani, A., Shazeer, N., Parmar, N., Uszkoreit, J., Jones, L., Gomez, A.N.,
  Kaiser, {\L}., Polosukhin, I.: Attention is all you need. In: Advances in
  neural information processing systems. pp. 5998--6008 (2017)

\bibitem{wang2018toward}
Wang, B., Zheng, H., Liang, X., Chen, Y., Lin, L., Yang, M.: Toward
  characteristic-preserving image-based virtual try-on network. In: Proceedings
  of the European Conference on Computer Vision (ECCV). pp. 589--604 (2018)

\bibitem{wang2021pyramid}
Wang, W., Xie, E., Li, X., Fan, D.P., Song, K., Liang, D., Lu, T., Luo, P.,
  Shao, L.: Pyramid vision transformer: A versatile backbone for dense
  prediction without convolutions. arXiv preprint arXiv:2102.12122  (2021)

\bibitem{wang2018non}
Wang, X., Girshick, R., Gupta, A., He, K.: Non-local neural networks. In:
  Proceedings of the IEEE conference on computer vision and pattern
  recognition. pp. 7794--7803 (2018)

\bibitem{wang2004image}
Wang, Z., Bovik, A.C., Sheikh, H.R., Simoncelli, E.P.: Image quality
  assessment: from error visibility to structural similarity. IEEE transactions
  on image processing  \textbf{13}(4),  600--612 (2004)

\bibitem{wei2018person}
Wei, L., Zhang, S., Gao, W., Tian, Q.: Person transfer gan to bridge domain gap
  for person re-identification. In: Proceedings of the IEEE conference on
  computer vision and pattern recognition. pp. 79--88 (2018)

\bibitem{xu2021drb}
Xu, W., Long, C., Wang, R., Wang, G.: Drb-gan: A dynamic resblock generative
  adversarial network for artistic style transfer. In: Proceedings of the
  IEEE/CVF International Conference on Computer Vision. pp. 6383--6392 (2021)

\bibitem{yang2018pose}
Yang, C., Wang, Z., Zhu, X., Huang, C., Shi, J., Lin, D.: Pose guided human
  video generation. In: Proceedings of the European Conference on Computer
  Vision (ECCV). pp. 201--216 (2018)

\bibitem{yang2021towards}
Yang, L., Wang, P., Liu, C., Gao, Z., Ren, P., Zhang, X., Wang, S., Ma, S.,
  Hua, X., Gao, W.: Towards fine-grained human pose transfer with detail
  replenishing network. IEEE Transactions on Image Processing  \textbf{30},
  2422--2435 (2021)

\bibitem{yin2020novel}
Yin, M., Sun, L., Li, Q.: Novel view synthesis on unpaired data by conditional
  deformable variational auto-encoder. In: European Conference on Computer
  Vision. pp. 87--103. Springer (2020)

\bibitem{zhang2021pise}
Zhang, J., Li, K., Lai, Y.K., Yang, J.: Pise: Person image synthesis and
  editing with decoupled gan. In: Proceedings of the IEEE/CVF Conference on
  Computer Vision and Pattern Recognition. pp. 7982--7990 (2021)

\bibitem{zhang2020cross}
Zhang, P., Zhang, B., Chen, D., Yuan, L., Wen, F.: Cross-domain correspondence
  learning for exemplar-based image translation. In: Proceedings of the
  IEEE/CVF Conference on Computer Vision and Pattern Recognition. pp.
  5143--5153 (2020)

\bibitem{zhang2018unreasonable}
Zhang, R., Isola, P., Efros, A.A., Shechtman, E., Wang, O.: The unreasonable
  effectiveness of deep features as a perceptual metric. In: Proceedings of the
  IEEE conference on computer vision and pattern recognition. pp. 586--595
  (2018)

\bibitem{zheng2021rethinking}
Zheng, S., Lu, J., Zhao, H., Zhu, X., Luo, Z., Wang, Y., Fu, Y., Feng, J.,
  Xiang, T., Torr, P.H., et~al.: Rethinking semantic segmentation from a
  sequence-to-sequence perspective with transformers. In: Proceedings of the
  IEEE/CVF Conference on Computer Vision and Pattern Recognition. pp.
  6881--6890 (2021)

\bibitem{zheng2017unlabeled}
Zheng, Z., Zheng, L., Yang, Y.: Unlabeled samples generated by gan improve the
  person re-identification baseline in vitro. In: Proceedings of the IEEE
  international conference on computer vision. pp. 3754--3762 (2017)

\bibitem{zhou2021cocosnet}
Zhou, X., Zhang, B., Zhang, T., Zhang, P., Bao, J., Chen, D., Zhang, Z., Wen,
  F.: Cocosnet v2: Full-resolution correspondence learning for image
  translation. In: Proceedings of the IEEE/CVF Conference on Computer Vision
  and Pattern Recognition. pp. 11465--11475 (2021)

\bibitem{zhu2020deformable}
Zhu, X., Su, W., Lu, L., Li, B., Wang, X., Dai, J.: Deformable detr: Deformable
  transformers for end-to-end object detection. arXiv preprint arXiv:2010.04159
  (2020)

\bibitem{zhu2019progressive}
Zhu, Z., Huang, T., Shi, B., Yu, M., Wang, B., Bai, X.: Progressive pose
  attention transfer for person image generation. In: Proceedings of the
  IEEE/CVF Conference on Computer Vision and Pattern Recognition. pp.
  2347--2356 (2019)

\end{thebibliography}

{

}

\clearpage
\appendix
{\LARGE\noindent\textbf{Appendix}}

\section{Network architectures}

In this section, we provide the details of network
structure. Table \ref{tab:tab2}, \ref{tab:tab3}, 
are the network structures of the encoder E, the generator G, 
respectively. In Conv and Residual Block, F, K and S respectively represent the output dimension, convolution kernel size and stride. IN and LN represent instance normalization and layer normalization, respectively.

\section{Comparisons with the state-of-the-arts}

\begin{figure*}[ht]
\centering
\includegraphics[width=1\textwidth]{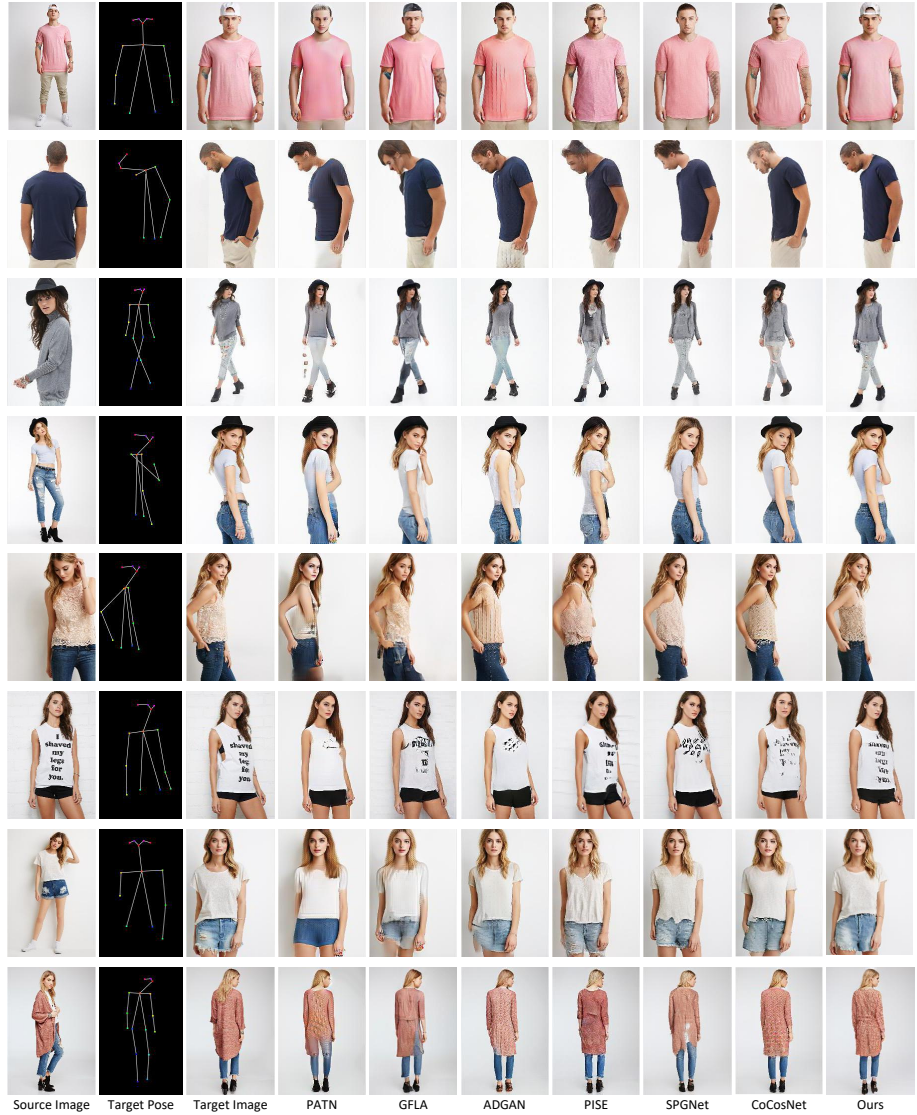}
\caption{Qualitative comparison between our method and other state-of-the-arts. The target ground truths and the synthesized results from each models are listed in rows.}
\vspace{0.5cm}
\label{fig:fig3}
\end{figure*}

In Fig \ref{fig:fig3}, We provide additional qualitative comparisons between our method and other state-of-the-arts(\emph{e.g.} PATN \cite{zhu2019progressive}, GFLA \cite{ren2020deep}, ADGAN \cite{men2020controllable}, PISE \cite{zhang2021pise}, SPGNet \cite{lv2021learning}, CoCosNet \cite{zhang2020cross}).
Results show that our method can generate more consistent appearance and pose with the target.

\section{Visualization of the generated parsing maps}

We also provide more visualization results of the generated parsing maps in Fig \ref{fig:fig4}. 
It is clear that cross attention matrix can  accurately predict the target parsing map regardless
of diverse pose and viewpoint changes, revealing the effectiveness of the proposed cross attention based style distribution module.

\section{Results of virtual try-on}

By exchanging the
channel feature of specific semantic region in the style features, our model can achieve virtual try-on task. Additional examples of virtual try-on are shown in Fig \ref{fig:fig5}.




\begin{table*}
\centering
\caption{The structure of encoder E. In E, we put $I_s^i$ into Pre-trained VGG19 network and take the features of the corresponding layers as side branches, then concat them together with the main branch. Note that we only show one source style $I_s^i$ as an example, where $i=1,2,\cdots,8$ is the semantic index. And all $I_s^i$ concat together lastly.}
\vspace{0.3cm}
\setlength{\tabcolsep}{12mm}
\renewcommand\arraystretch{1.3}
\begin{tabular}{|c|c|} 
\hline
\textbf{Input}                                                                                   & $I_s^i$ $(256\times176\times3)$                                                                                             \\ 
\hline
\multirow{10}{*}{\begin{tabular}[c]{@{}c@{}}\textbf{Intermediate}\\\textbf{ Layers}\end{tabular}} & $\rm{Conv(F=64, K=7, S=1), ReLU}$                                                                                           \\ 
\cline{2-2}
                                                                                                 & $\rm {Concat(Pre-trained\ VGG19\   conv1\_1)} $        \\ 
\cline{2-2}
                                                                                                 & $\rm{Conv(F=128, K=4, S=2), ReLU}$                                                                                          \\ 
\cline{2-2}
                                                                                                & $\rm {Concat(Pre-trained\ VGG19\   conv2\_1)} $        \\  
\cline{2-2}
                                                                                                 & $\rm{Conv(F=256, K=4, S=2), ReLU}$                                                                                          \\ 
\cline{2-2}
                                                                                                & $\rm {Concat(Pre-trained\ VGG19\   conv3\_1)} $        \\ 
\cline{2-2}
                                                                                                 & $\rm{Conv(F=512, K=4, S=2), ReLU}$                                                                                          \\ 
\cline{2-2}
                                                                                               & $\rm {Concat(Pre-trained\ VGG19\   conv4\_1)} $        \\ 
\cline{2-2}
                                                                                                 & Avg Pooling  \\
                                                    \cline{2-2}   
                                                                                               
                                   & $\rm{Conv(F=256, K=1, S=1)}$                                                             \\ 
\hline
\textbf{Output}                                                                                  & $F_s^i$ $(1\times1\times256)$                                                                                               \\
\hline
\end{tabular}

\label{tab:tab2}
\end{table*}

\begin{table}[ht]
\centering
\caption{The structure of the generator G. In AdaIN ResBlocks and AFN ResBlocks, the content in bracket is used as side branch to affect the main branch. }
\vspace{0.3cm}
\setlength{\tabcolsep}{0.65mm}
\renewcommand\arraystretch{1.3}
\begin{tabular}{|c|c|c|} 
\hline
\textbf{Input}                                                                                   & $P_t$ $(256\times176\times30)$                        & $F_s$ $(1\times1\times2048)$    \\ 
\hline
\multirow{14}{*}{\begin{tabular}[c]{@{}c@{}}\textbf{Intermediate}\\\textbf{Layers}\end{tabular}} & $\rm{Conv(F=64, K=7, S=1), IN, ReLU}$                 & $\rm{Fc(2048), ReLU}$           \\
                                                                                                 & $\rm{Conv(F=128, K=4, S=2), IN, ReLU}$                & $\rm{Fc(256), ReLU}$            \\
                                                                                                 & $\rm{Conv(F=256, K=4, S=2), IN, ReLU}$                & $\rm{Fc(256), ReLU}$            \\
                                                                                                 & $\rm{F_{p} = Residual\  Blocks(F=256, K=3, S=1)\times8}$ & $\rm{F_{s'} = Fc(8192), ReLU}$  \\ 
\cline{2-3}
                                                                                                 & \multicolumn{2}{c|}{$\rm{F_{crs}=AdaIN\  ResBlock(F_{s'})}$}                              \\ 
\cline{2-3}
                                                                                                 & \multicolumn{2}{c|}{$\rm{F_{ps}=CASD\  (F_{crs}, F_p, F_s)}$}                   \\
                                                                                                 & \multicolumn{2}{c|}{$\rm{F_{ps}=CASD\  (F_{ps}, F_p, F_s)}$}                    \\ 
\cline{2-3}
                                                                                                 & \multicolumn{2}{c|}{$\rm{F_{p'}=AFN\  ResBlocks(F_{ps})}$}                                 \\
\cline{2-3}                                                                                  & \multicolumn{2}{c|}{$\rm{UpSample(scale\_factor=2)}$}                                   \\
                                                                                                 & \multicolumn{2}{c|}{$\rm{Conv(F=128, K=5, S=1), LN, ReLU}$}                             \\
                                                                                                 & \multicolumn{2}{c|}{$\rm{UpSample(scale\_factor=2)}$}                                   \\
                                                                                                 & \multicolumn{2}{c|}{$\rm{Conv(F=64, K=5, S=1), LN, ReLU}$}                             \\
                                                                                                 & \multicolumn{2}{c|}{$\rm{Conv(F=3, K=7, S=1), Tanh}$}                                  \\ 
\hline
\textbf{Output}                                                                                  &                                                              \multicolumn{2}{c|}{$\hat{I}_t\                                  (256\times176\times3)$}                                                        \\
\hline
\end{tabular}
\label{tab:tab3}
\end{table}


\begin{figure*}[ht]
\centering
\includegraphics[width=1\textwidth]{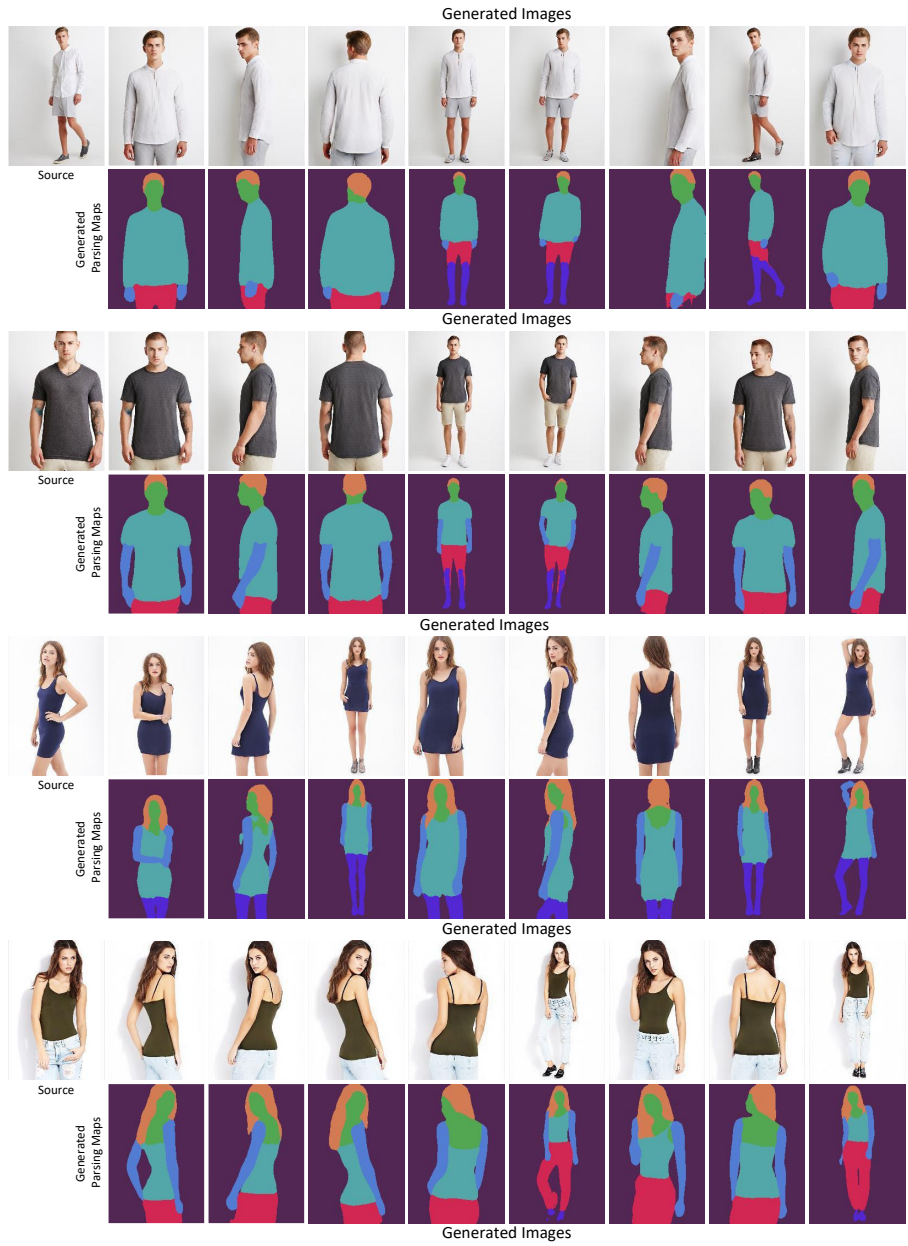}
\caption{Given the source image, our model is able to transfer the pose as required. The synthesized person and visualization of the generated target parsing maps are shown. }
\vspace{0.2cm}
\label{fig:fig4}
\end{figure*}

\begin{figure*}[ht]
\centering
\includegraphics[width=0.98\textwidth]{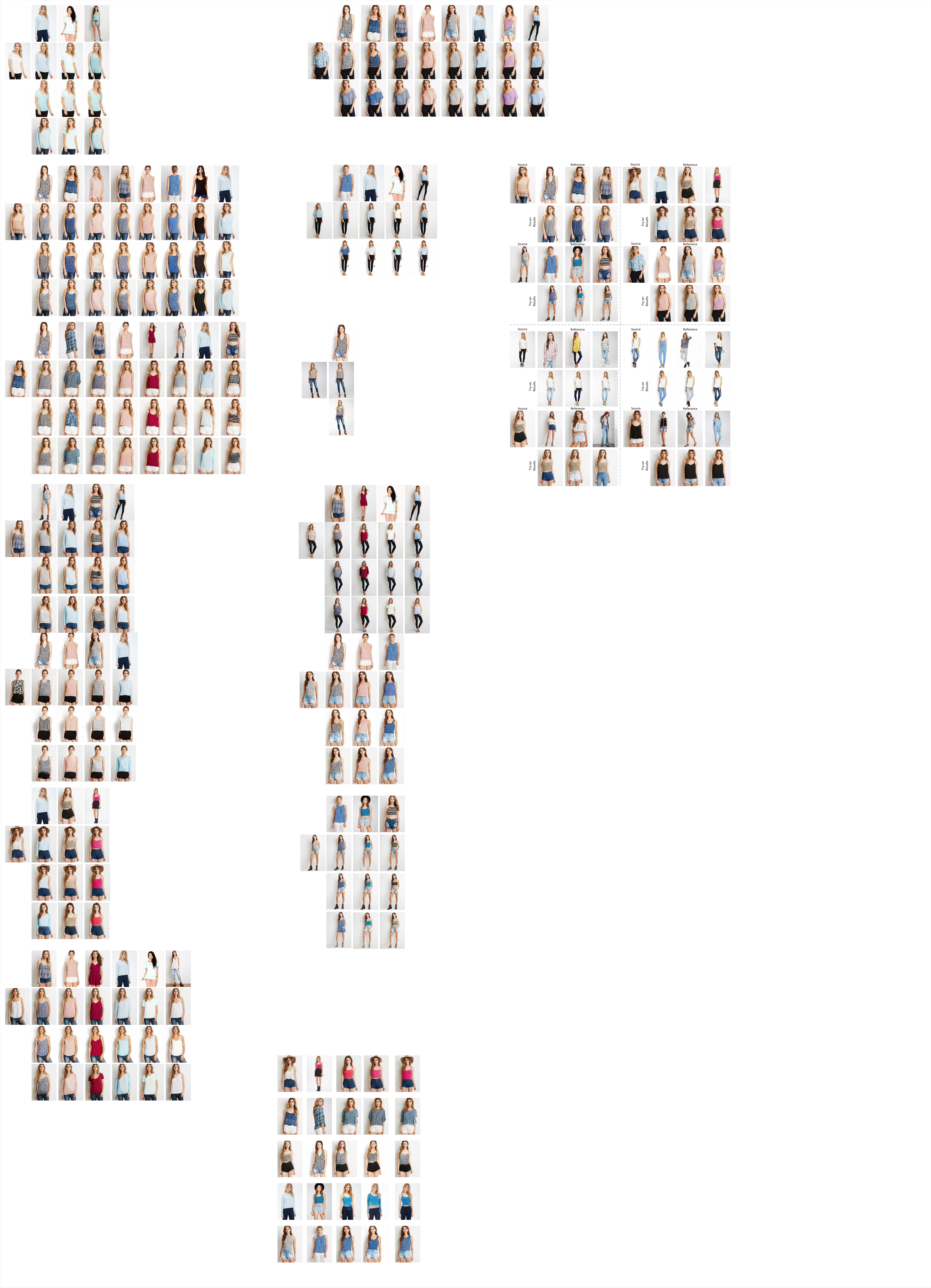}
\caption{Given the source image and reference images, our model is able to perform virtual try-on task. The top half is the results of trying on the upper-clothes and the bottom half is the results of trying on the pants.}
\vspace{0.2cm}
\label{fig:fig5}
\end{figure*}

\end{document}